\pdfoutput=1
\documentclass[runningheads]{llncs}

\usepackage[year=2026]{eccv}

\usepackage{eccvabbrv}

\usepackage{booktabs}
\usepackage{graphicx}
\usepackage{subcaption}
\usepackage{booktabs}      
\usepackage{multirow}      
\usepackage[table]{xcolor} 
\usepackage{colortbl}
\usepackage{wrapfig}       

\usepackage[accsupp]{axessibility}  

\usepackage[breaklinks,colorlinks,citecolor=eccvblue]{hyperref}

\usepackage{orcidlink}

\begin{document}

\title{Zip-GS: Extreme Compression of 3D Gaussians with Diffusion Priors} 

\titlerunning{Zip-GS}

\author{
Jiaqi Chen\textsuperscript{1,2,*} \and
Xinhao Ji\textsuperscript{2,*} \and
Yuanyuan Gao\textsuperscript{1,2,*} \and
Hao Li\textsuperscript{1} \and
Yuning Gong\textsuperscript{2} \and
Yifei Liu\textsuperscript{2} \and
Dingwen Zhang\textsuperscript{1,$\dagger$} \and
Dan Xu\textsuperscript{4} \and
Xiao Sun\textsuperscript{2} \and
Zhihang Zhong\textsuperscript{3,$\dagger$}
}

\authorrunning{J. Chen et al.}

\institute{
Northwestern Polytechnical University \\
\and
Shanghai Artificial Intelligence Laboratory
\and
Shanghai Jiao Tong University
\and
Hong Kong University of Science and Technology
}

\maketitle
\renewcommand{\thefootnote}{}
\footnotemark
\footnotetext{* denotes equal contribution. This work was done during their internship at Shanghai Artificial Intelligence Laboratory. $\dagger$ denotes corresponding author.}

\begin{figure}[h]
    \centering
    \includegraphics[width=1.0\textwidth]{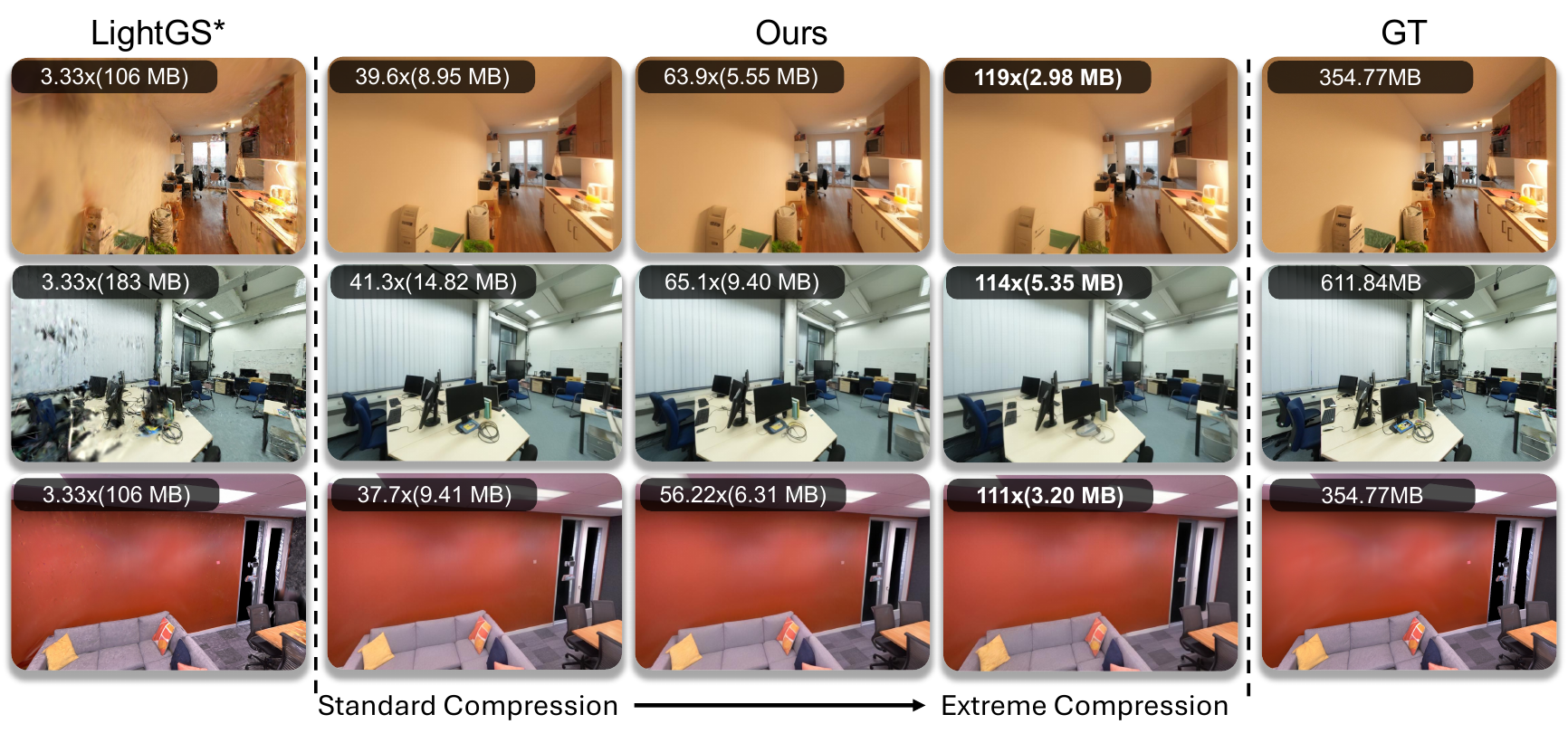}
    \caption{\textbf{Qualitative comparison under different compression ratios.}
    LightGS* indicates the optimized version of LightGS\cite{fan2024lightgaussian}. LightGS* achieves only limited compression (around $3\times$) and produces severe visual artifacts. 
    In contrast, our method realizes orders-of-magnitude higher compression (exceeding $100\times$ in some cases) while preserving scene geometry and appearance realism. 
    Even at aggressive compression ratios, our reconstructions remain close to ground-truth quality, demonstrating the robustness of the proposed framework. }
    \label{fig:teaser}
\end{figure}

\begin{abstract}
As world models increasingly produce Internet-scale 3D content, 
efficient delivery of neural assets becomes a critical systems bottleneck. 
Neural scene representations, especially 3D Gaussian Splatting (3DGS), 
provide excellent rendering quality and speed, but their memory footprint remains too large for bandwidth-limited storage and streaming. 
Prior 3DGS compression pipelines face a persistent trade-off: optimization-based methods, which require ground-truth images and per-scene fine-tuning, 
are accurate yet slow and often inapplicable to Internet-scale 3D assets without paired supervision, 
while lightweight pruning/quantization schemes are efficient but fragile at very high compression levels. 
To address this gap, we present ZipGS, a fully feed-forward framework for extreme 3DGS compression and restoration. 
ZipGS combines Universal Gaussian Compression (UGC) and GaussPainter in a two-stage design. 
UGC removes redundant Gaussians without re-optimization, 
keeping a compact set of structurally important primitives. 
GaussPainter then recovers visual quality from severely compressed renderings via mask-guided diffusion priors, while also refining visible regions rather than only completing missing areas. 
With a lightweight VAE and one-step diffusion sampling, our method supports practical real-time restoration. Across indoor and outdoor benchmarks, 
ZipGS consistently improves perceptual and distortion metrics over strong compression and generative baselines, achieving compression beyond $100\times$ and up to $300\times$ in challenging settings. These results demonstrate that diffusion priors can effectively reconcile extreme compression with high-fidelity neural rendering.
\end{abstract}

\section{Introduction}
\label{sec:intro}
In recent years, rapid progress in world models and generative 3D content creation has enabled the Internet-scale sharing of 3D assets, often without explicit per-view ground-truth supervision. These assets are quickly becoming core infrastructure for immersive content ecosystems and the emerging 3D Internet, underpinning applications such as augmented/virtual reality, interactive visualization, remote collaboration, and digital twin systems. As a result, efficient storage, transmission, and streaming of high-quality 3D representations is critical, yet they remain bulky and expensive, posing a major obstacle in bandwidth-constrained or mobile scenarios.

Among recent representations, 3D Gaussian Splatting (3DGS)~\cite{kerbl3Dgaussians} has emerged as a compelling choice for real-time neural rendering and novel view synthesis, explicitly representing scenes with millions of anisotropic Gaussian primitives rendered efficiently in screen space. However, the sheer number of Gaussians still leads to substantial storage and communication overhead, with per-scene sizes reaching hundreds of megabytes or even gigabytes, which becomes a major bottleneck for Internet-scale 3D asset transmission.

To resolve the redundancy in Gaussian primitives, existing methods can be broadly categorized into two groups: optimization-based approaches~\cite{niedermayr2024compressed,liu2025maskgaussian} that fine-tune Gaussian parameters, and training-free approaches~\cite{fan2024lightgaussian,tian2025flexgaussian} that rely on quantization or heuristic pruning. Optimization-based pipelines typically assume access to multi-view ground-truth images and perform costly per-scene optimization, which limits their applicability to Internet-scale 3D assets synthesized by generative models, where only the neural representation is available and no paired images exist. Training-free methods are lightweight and scalable but struggle under aggressive compression ratios, leading to degraded rendering quality and a persistent trade-off between compression ratio and fidelity.


Motivated by this challenge, we propose ZipGS, an extreme 3DGS compression framework tailored for large-scale 3D asset delivery. ZipGS leverages generative priors to push compression to very high ratios while maintaining strong visual fidelity at render time, enabling practical storage and transmission of 3DGS assets over the Internet. Concretely, ZipGS consists of two complementary modules: Universal Gaussian Compression (UGC) and GaussPainter. UGC performs extreme compression without per-scene re-optimization, retaining only essential information and maximizing the benefit of generative priors. GaussPainter then exploits the strong priors of diffusion models with a mask-guided refinement strategy to restore high-quality renderings from heavily pruned and incomplete Gaussian scenes.

By combining UGC and GaussPainter, our framework achieves orders of magnitude compression, often exceeding $100\times$, while still enabling real-time rendering and high-quality restoration. In contrast, our method maintains geometric fidelity and perceptual realism even under much stronger compression, demonstrating its robustness and clear advantage over existing baselines.
Our contributions are threefold:  

(1) We demonstrate, for the first time, that a data-driven paradigm can drive generative 3DGS compression, moving beyond heuristic pruning or optimization-based tuning by leveraging powerful generative priors for faithful restoration under extreme compression.

(2) We propose two complementary modules: UGC for re-optimization-free pruning and GaussPainter for diffusion-based refinement, and demonstrate that their synergy enables compact representations while preserving high visual fidelity.

(3) Extensive experiments on public datasets demonstrate that ZipGS achieves state-of-the-art performance, maintaining real-time rendering and robust quality even under a storage size compression ratio of up to $100\times$.

\section{Related Work}
\subsection{Efficient Neural Rendering.}  
3D Gaussian Splatting (3DGS) has rapidly emerged as a powerful representation for real-time scene rendering, yet its efficiency is often constrained by the large number of Gaussian primitives. To improve scalability, prior works have explored compression and pruning strategies. Codebook-based methods~\cite{navaneet2024compgs,lee2024compact,fan2024lightgaussian} reduce the storage footprint by quantizing Gaussian parameters, while pruning approaches~\cite{fan2024lightgaussian,tian2025flexgaussian} directly eliminate redundant primitives to achieve compact representations. Beyond pure reduction, densification techniques such as Mini-Splatting~\cite{fang2024mini} and Taming-3DGS~\cite{mallick2024taming} regenerate new primitives in a spatially efficient manner, whereas Scaffold-GS~\cite{lu2024scaffold} anchors Gaussians on structured grids for better distribution.  

Although these approaches have demonstrated strong effectiveness, they often rely on retraining or fixed pruning ratios, limiting flexibility under different compression requirements. In contrast, our method improves upon the LightGS~\cite{fan2024lightgaussian} scoring mechanism and further integrates generative priors, enabling controllable compression while preserving rendering quality even at extreme ratios.

\subsection{Diffusion Priors for 3D Reconstruction.}  
Diffusion models have recently shown strong potential in addressing degradation in 3D reconstruction tasks. Traditional dense methods, such as NeRF-based pipelines, require abundant multi-view inputs, while sparse-view methods~\cite{wang2023sparsenerf,chen2023structnerf} attempt to compensate with structural priors but still suffer from incomplete reconstructions. Diffusion models extend this line of work by providing powerful generative priors that can restore missing details and refine degraded regions.  

Several recent works integrate diffusion with Gaussian-based or NeRF-based reconstructions. For instance, Difix3D+~\cite{wu2025difix3d+} leverages diffusion to denoise target views using reference information, yielding improved fidelity. Generative Gaussian Splatting (GGS)~\cite{schwarz2025generative} integrates video diffusion priors with Gaussian splatting to enhance view consistency, and GSFixer~\cite{yin2025gsfixer} employs reference-guided diffusion to correct artifacts in sparse inputs. Similarly, GSD~\cite{mu2024gsdiffusion} combines Gaussian splatting and diffusion priors for single-view 3D reconstruction, while Single-Stage Diffusion NeRF (SSDNeRF)~\cite{chen2023single} jointly learns diffusion priors and neural fields from sparse inputs. Diffusion has also been explored for human reconstruction, e.g., DiHuR~\cite{chen2024dihur}, which uses diffusion priors to enhance geometry and texture fidelity under sparse or degraded observations.  

Different from these approaches, our method directly applies diffusion priors to compressed Gaussian renderings, without requiring extra views or heavy supervision. By combining mask-guided conditioning with efficient one-step diffusion and latent alignment, it is capable of handling diverse degradation sources such as sparse inputs, severe pruning, and structural inconsistencies, thereby producing more consistent and high-quality reconstructions.
\section{Preliminaries}

\subsection{3D Gaussian Splatting}

3D Gaussian Splatting \cite{kerbl3Dgaussians} explicitly represents 3D scenes with a collection of anisotropic Gaussian primitives. Each Gaussian is parameterized by its mean position $\mu \in \mathbb{R}^3$, covariance matrix $\Sigma \in \mathbb{R}^{3\times 3}$, opacity $\sigma \in (0,1]$, and color (or feature) $\mathbf{c} \in [0,1]^3$.  
During rendering, the color $C$ of a pixel is obtained by blending $N$ ordered Gaussians that overlap with the pixel using alpha blending:

\begin{equation}
C = \sum_{i \in N} \mathbf{c}_i \alpha_i 
\prod_{j=1}^{i-1} (1 - \alpha_j),
\label{eq:3dgs-blend}
\end{equation}

where $\alpha_i$ denotes the opacity contribution of the $i$-th Gaussian at the pixel, typically computed by evaluating the projected 2D elliptical footprint defined by $\Sigma_i$ and scaling with its opacity $\sigma_i$.  
This design enables differentiable and efficient rendering in real time, making 3DGS a strong alternative to volumetric ray marching methods.
\subsection{Diffusion Models}

Diffusion models \cite{ho2020denoising,song2020score} generate data by adding Gaussian noise in a forward Markov chain and learning to reverse it.  
In the forward process, noise is added step by step:

\begin{equation}
q(\mathbf{z}_{1:T} \mid \mathbf{z}_0) = \prod_{t=1}^{T} 
\mathcal{N}\!\left(\mathbf{z}_t;\, \sqrt{1-\beta_t}\,\mathbf{z}_{t-1},\, \beta_t \mathbf{I}\right),
\label{eq:forward-step}
\end{equation}

which has the closed form
\begin{equation}
q(\mathbf{z}_t \mid \mathbf{z}_0) = 
\mathcal{N}\!\left(\mathbf{z}_t;\, \sqrt{\bar{\alpha}_t}\,\mathbf{z}_0,\,
(1-\bar{\alpha}_t)\mathbf{I}\right),
\end{equation}
where $\alpha_t = 1-\beta_t$ and $\bar{\alpha}_t = \prod_{i=1}^t \alpha_i$.

\subsubsection{Reverse process.}
Sampling starts from Gaussian noise $\mathbf{z}_T \sim \mathcal{N}(0,I)$ and reconstructs $\mathbf{z}_0$ via

\begin{equation}
q(\mathbf{z}_{t-1}\mid \mathbf{z}_t, \mathbf{z}_0) =
\mathcal{N}\!\left(\mathbf{z}_{t-1};\, \mu_t(\mathbf{z}_t,\mathbf{z}_0),\, \sigma_t^2 \mathbf{I}\right),
\end{equation}

with mean
\begin{equation}
\mu_t(\mathbf{z}_t,\mathbf{z}_0) =
\frac{1}{\sqrt{\alpha_t}}\Big(\mathbf{z}_t - 
\frac{1-\alpha_t}{\sqrt{1-\bar{\alpha}_t}}\,\boldsymbol{\epsilon}\Big),
\end{equation}
where $\boldsymbol{\epsilon}$ is the injected noise.  
A neural network $\boldsymbol{\epsilon}_\theta(\mathbf{z}_t,t,\mathbf{c})$ predicts $\boldsymbol{\epsilon}$, optionally conditioned on $\mathbf{c}$.  

The training objective is the noise-prediction loss:
\begin{equation}
\mathcal{L}_{\text{DM}} =
\mathbb{E}_{\mathbf{z}_0,t,\boldsymbol{\epsilon}}\,
\left\|\boldsymbol{\epsilon} - 
\boldsymbol{\epsilon}_\theta\!\left(\sqrt{\bar{\alpha}_t}\mathbf{z}_0 +
\sqrt{1-\bar{\alpha}_t}\,\boldsymbol{\epsilon},t\right)\right\|_2^2.
\end{equation}

\section{Leveraging generative model priors for extreme Gaussian compression}

Recent feed-forward world models and reconstruction systems can produce large volumes of 3D assets, but these assets are often too heavy to be stored, transmitted, and shared efficiently at scale. Our goal is to establish a practical compression paradigm for 3D assets that makes them truly deliverable over the Internet by leveraging strong generative priors.

\begin{figure}[t]
  \centering
  \includegraphics[width=0.9\linewidth]{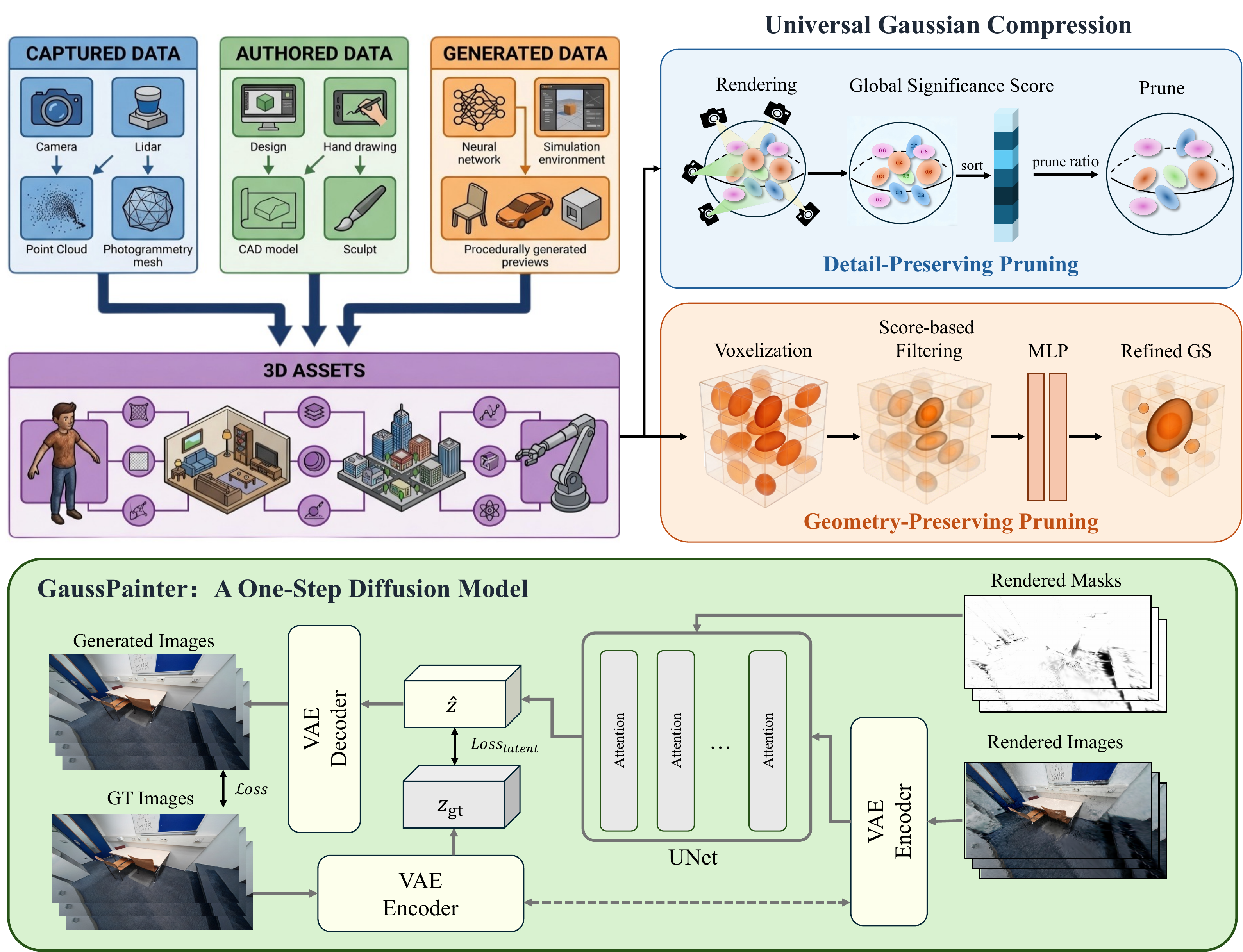}
  \caption{\textbf{Overview of the proposed Zip-GS framework.} The pipeline contains two stages: Universal Gaussian Compression (UGC), which aggressively removes redundancy while preserving key geometric and appearance cues, and GaussPainter, which restores high-fidelity renderings from compressed Gaussian scenes using mask-guided diffusion priors.}
  \label{fig:framework_zipgs}
\end{figure}

To this end, we propose a two-stage framework, illustrated in Fig.~\ref{fig:framework_zipgs}. In the first stage, we exploit the structural properties of 3D Gaussian representations to derive an extremely compact, transmission-friendly asset file (Sec.~\ref{sec:ugc}). In the second stage, we render the compressed asset efficiently and enhance the resulting images using generative priors, compensating for the information lost during extreme compression (Sec.~\ref{sec:gausspainter}).

\subsection{Universal Gaussian Compression}
\label{sec:ugc}

Given a 3DGS asset without ground-truth images, our compression method aims to produce a representation with as few Gaussians as possible while preserving the information most beneficial for subsequent diffusion-based generation.
Because the generative model must recover missing information, our compression method prunes Gaussians along two complementary axes to reduce the count while preserving scene content and minimizing hallucinations: Detail-Preserving Pruning (DPP), which preserves fine rendering details, and Geometry-Preserving Pruning (GPP), which retains the overall geometric structure.

\subsubsection{Detail-Preserving Pruning.}
To quantify the contribution of each Gaussian primitive and guide pruning, we follow LightGS~\cite{fan2024lightgaussian} and define a global significance score $GS_j$ that accounts for visibility, opacity, and ray transmittance:

\begin{equation}
GS_j = \sum_{i=1}^{MHW} \mathbf{1}\big(G(\mathbf{X}_j), r_i\big)\,\cdot\, \sigma_j \,\cdot\, T_j,
\label{eq:gs}
\end{equation}

where $i$ indexes all pixels across $M$ views with image resolution $H \times W$. The indicator function $\mathbf{1}(\cdot)$ returns $1$ if the $j$-th Gaussian $G(\mathbf{X}_j)$ intersects the camera ray $r_i$, and $0$ otherwise. $\sigma_j$ denotes the opacity of the $j$-th Gaussian, and $T_j$ represents the accumulated transmittance of the ray before reaching $G(\mathbf{X}_j)$, defined as:

\begin{equation}
T_j = \prod_{i=1}^{j-1} (1 - \sigma_i).
\label{eq:transmittance}
\end{equation}

In a Gaussian-based 3D asset, feed-forward reconstruction often converts many pixels into primitives, producing substantial redundancy with limited impact on rendering quality. In DPP, our goal is to reduce the Gaussian count while retaining fine appearance details and view-consistent textures. We therefore compute $GS_j$ for each Gaussian using either randomly sampled or provided camera poses, rank the primitives by their scores, and prune those with low importance. This scoring-based pruning preserves visually salient structures while removing redundant primitives, providing a compact yet detail-preserving input for the subsequent generative refinement.

\begin{wrapfigure}{r}{0.48\textwidth}
  \vspace{-10pt}
  \centering
  \includegraphics[width=\linewidth]{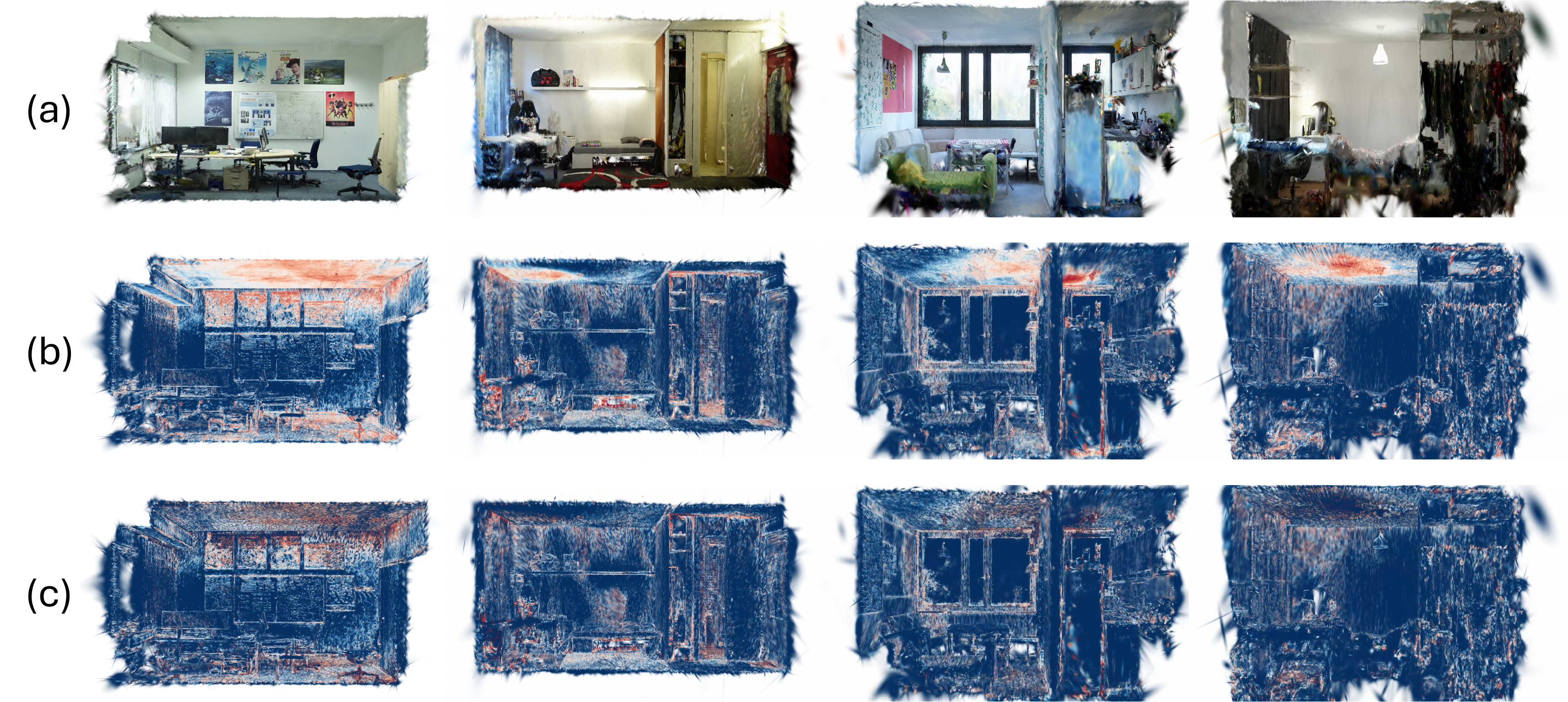}
  \caption{\textbf{Detail-preserving scoring and geometry-aware correction.} (a) Rendered views of Gaussian scenes. (b) Heatmaps of the global significance score used in DPP, which emphasize fine textures but under-cover sparse regions. (c) The voxel-based supplementation in GPP rebalances the distribution, yielding more uniform coverage and improved preservation of structure and details.}
  \vspace{-0.4cm}
  \label{fig:global_significance_scoring}
\end{wrapfigure}

\subsubsection{Geometry-Preserving Pruning.}
We observe that the importance score is often influenced by the viewing direction and the geometric position of Gaussians, as illustrated in Fig.~\ref{fig:global_significance_scoring}. While it preserves color details, for complete 3D assets it can severely damage geometry, leading to holes in rendered images. To better leverage diffusion priors, we introduce Geometry-Preserving Pruning, which complements the scoring function. With the same number of Gaussians, this sampling approach more completely preserves both global geometric structures and fine-grained local details.

The significance score does not ensure spatial uniformity: dense regions may retain redundancy, while sparse ones may lose coverage. We partition the space into voxels of size $v$, group Gaussians by voxel, and ensure each voxel with sufficient points contributes at least one selected Gaussian. For a voxel $\mathcal{V}$ containing Gaussians $\{g_j\}$ with corresponding 
significance scores $\{s_j\}$, we select $k$ Gaussians, denoted by 
$\mathcal{S}_{\mathcal{V}}$, that maximize the total significance score 
within the voxel:
\begin{equation}
\mathcal{S}_\mathcal{V} = \arg\max_{\substack{\mathcal{S} \subset \mathcal{V} \\ |\mathcal{S}| = k}} \sum_{g_j \in \mathcal{S}} s_j,
\end{equation}

where $k = \lfloor |\mathcal{V}| \cdot \rho \rfloor$. The final set combines globally selected Gaussians with voxel-based supplements, balancing global importance and spatial coverage.

However, the Gaussians supplemented by this strategy are often small; although geometric details are retained, these Gaussians may occupy only tiny regions in the rendered image and thus provide insufficient cues for the diffusion model. To address this, we add an MLP to predict the scales of the Gaussians retained after Geometry-Preserving Pruning. This MLP is trained jointly with the diffusion model, enabling the Gaussians to adapt their scales to parameters that are most beneficial for diffusion-based completion.

\subsection{\textbf{GaussPainter}: Efficient Mask-Guided Diffusion for Gaussian Scenes}
\label{sec:gausspainter}
The generative module is designed to leverage the strong priors of diffusion models to address the information loss introduced by aggressive Gaussian compression. While diffusion excels at synthesizing realistic content, it often suffers from two limitations in this setting: (i) hallucination of non-existent structures, and (ii) prohibitive time costs due to iterative sampling. 


\subsubsection{Mitigating Hallucinations.}

\begin{wrapfigure}{r}{0.48\textwidth}
  \centering
  \includegraphics[width=\linewidth]{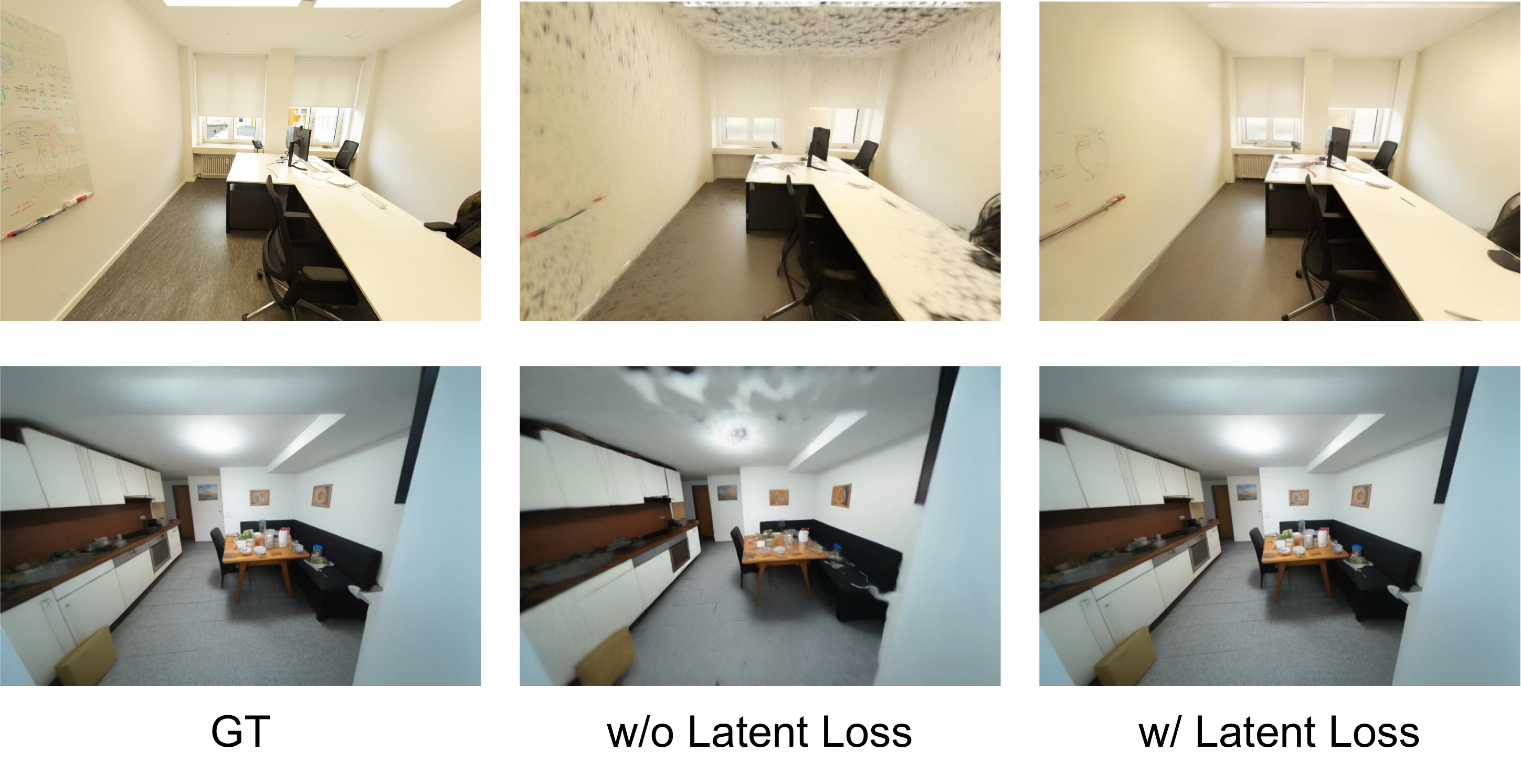}
  \caption{\textbf{Comparison of pixel-level supervision and latent supervision.} 
  From left to right: ground-truth image, compressed rendering, pixel-level supervision, and latent supervision. 
  Latent supervision better distinguishes regions to preserve from those to complete, reducing artifacts in heavily degraded areas.}
  \label{fig:comparison_latent_loss}
  \vspace{-0.5cm}
\end{wrapfigure}

To mitigate hallucination in compressed Gaussian renderings, we employ two complementary strategies: \textbf{latent supervision} and \textbf{mask guidance}. For \textbf{latent supervision}, relying only on pixel\text{-}level losses fails to reliably separate regions that should be preserved from those requiring completion, as corrupted or blackened areas introduced by Gaussian compression are often misread as valid content (see Fig.~\ref{fig:comparison_latent_loss}). Inspired by deblurring models~\cite{liu2025one}, we therefore impose supervision directly in the latent space: the ground\text{-}truth image $y$ and the degraded rendering $x$ are encoded by the VAE as $z_{\mathrm{hq}}{=}E(y)$ and $z_{\mathrm{lq}}{=}E(x)$, and we minimize
\[
\mathcal{L}_{\text{latent}}=\|z_{\mathrm{lq}}-z_{\mathrm{hq}}\|_2^2,
\]
encouraging the degraded latent to approach the manifold of its high\text{-}quality counterpart and helping the model decide what to preserve versus plausibly complete. 

\begin{wrapfigure}{r}{0.48\textwidth}
  \vspace{+0.2cm}
  \centering
  \includegraphics[width=\linewidth]{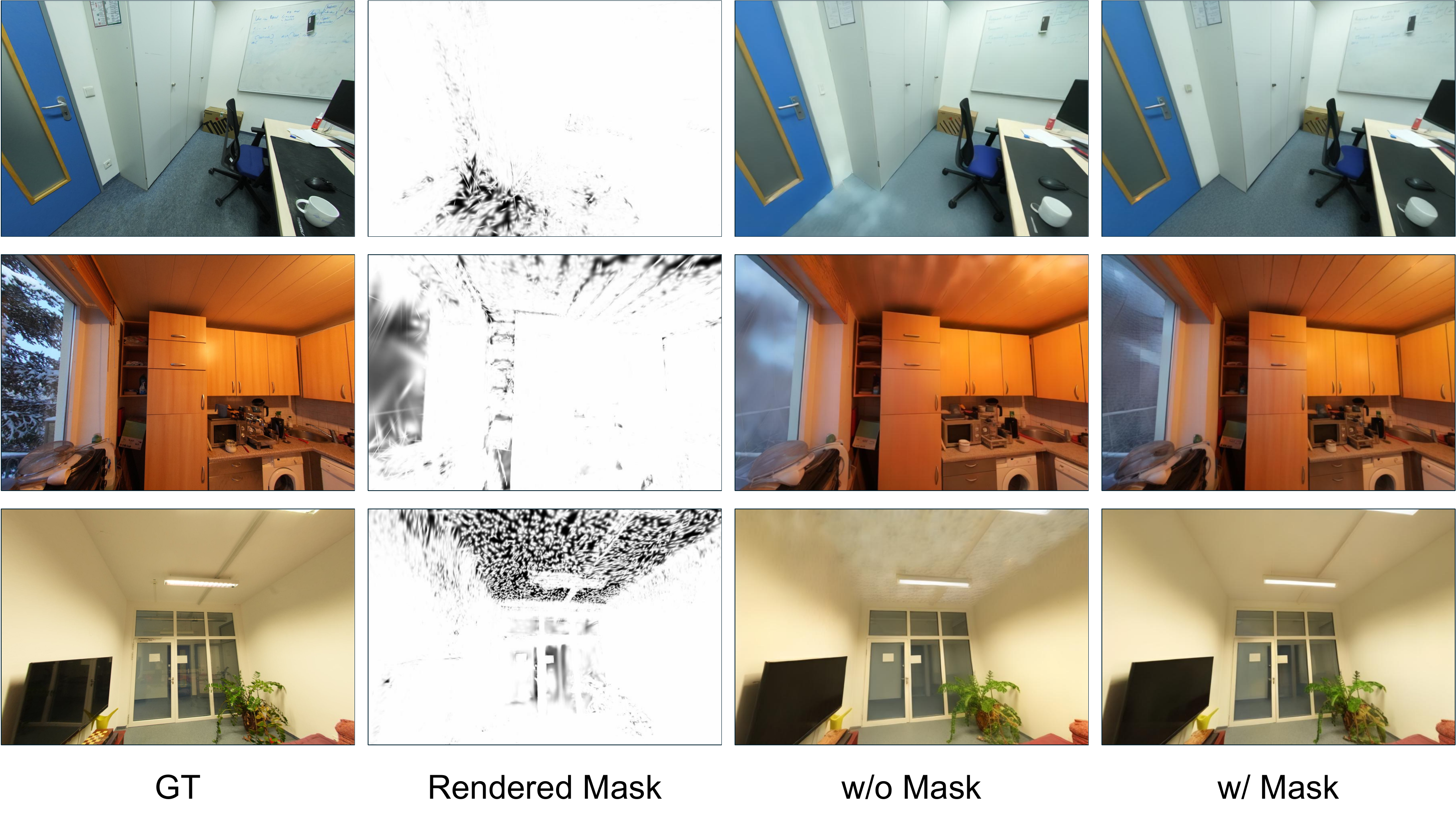}
  \caption{
  \textbf{Comparison of diffusion generation with and without mask guidance.} 
  From left to right: ground-truth image, rendered opacity mask from 3DGS blending, 
  generation without mask guidance, and generation with mask guidance. 
  The mask effectively constrains the diffusion model to complete missing regions while preserving visible structures, 
  leading to more faithful and realistic reconstructions.
  }
  \label{fig:mask_guided_results}
  \vspace{-0.5cm}
\end{wrapfigure}

While this latent alignment improves structural completion, large holes or low\text{-}texture regions may still exhibit color drift (see Fig.~\ref{fig:mask_guided_results}). To address this, we integrate \textbf{mask guidance}. A visibility mask derived from accumulated opacity in 3DGS rendering (Eq.~\ref{eq:3dgs-blend}) is normalized, flattened, globally pooled, and projected into the text\text{-}conditioning embedding space. The resulting mask embedding is then added element\text{-}wise to the caption features, which down\text{-}weights already reliable areas and steers the denoiser toward pixels likely to be missing or corrupted. This improves boundary handling around holes and reduces the tendency to propagate artifacts originating from compression.
At inference, the latent of the incomplete rendering is denoised at a fixed timestep $t{=}199$ in a single forward pass of the U\text{-}Net, where the injected mask embedding supplies spatial cues for where to inpaint versus preserve. This lightweight integration adds negligible overhead yet substantially reduces hallucinations and stabilizes color and illumination. Consistent with our overall design of directly applying diffusion priors to compressed Gaussian renderings without extra views or heavy supervision, combining mask\text{-}guided conditioning with efficient one\text{-}step diffusion and latent alignment enables GaussPainter to handle diverse degradation sources such as severe pruning, sparse inputs, and structural inconsistencies, thereby producing reconstructions that are structurally complete and visually faithful.

\subsubsection{Real-Time Rendering.}

A key advantage of our framework is its ability to operate in near real-time, making it suitable for practical deployment in interactive applications. To achieve this, we replace the standard VAE with TAESD~\cite{taesd}, a lightweight autoencoder designed for fast encoding and decoding while maintaining visual fidelity. In addition, we adopt a one-step diffusion schedule, where the denoiser directly predicts the clean latent in a single forward pass at a fixed timestep, bypassing the expensive iterative sampling typical of diffusion models. Together, these two design choices drastically reduce inference latency, enabling high-quality restoration of compressed Gaussian renderings at interactive frame rates.

For efficiency, we measure inference speed on an NVIDIA A100 GPU. After two warm-up iterations, we run 10 forward passes at $512^2$ resolution and report the mean runtime and standard deviation (Tab.~\ref{tab:inference_time}). Our method requires 2.13\,s for compression and 0.23\,s for decompression, demonstrating favorable computational cost compared to optimization-based approaches.

\section{Experiment}
\subsection{Experimental Setups}
\subsubsection{Training Strategy and Data.}
We initialize our model from the official pretrained weights of SD-Turbo and TASED.
During training, we fine-tune the full U-Net while applying a LoRA module (rank=4)
to the VAE decoder for parameter-efficient adaptation.
Training is conducted on DL3DV and the official training split of ScanNet++.
The DL3DV set includes all available video sequences.
\subsubsection{Evaluation Protocol.}  
We evaluate our method on 50 official ScanNet++ test scenes~\cite{yeshwanthliu2023scannetpp}. 
Across multiple compression ratios, we report PSNR, SSIM, LPIPS, 
together with the corresponding storage size.
To assess out-of-domain generalization, 
we further conduct experiments on three unseen datasets: 
Replica~\cite{replica19arxiv}, 
Mip-NeRF360~\cite{barron2021mip}, 
and Tanks \& Temples~\cite{Knapitsch2017}. 

\subsubsection{Baselines.} 
We compare our framework with both compression-based and generative methods. For compression, we include LightGaussian (LightGS)~\cite{fan2024lightgaussian} and FlexGaussian~\cite{tian2025flexgaussian}. For fair comparison, all compression baselines are evaluated without access to ground-truth images.
Accordingly, we adopt the GT-free variant of LightGS.
Moreover, our experiments are conducted under substantially more aggressive compression ratios than those commonly reported in prior work, resulting in a significantly more challenging evaluation setting. Additionally, we remove high-order SH components in our method for more extreme compression; accordingly, we report results for LightGS both with and without SH. 
In addition, we evaluate the original optimization-based LightGS by running its official optimization routine for 1,000 steps on the random views. For generative baselines, we evaluate PrefPaint~\cite{liu2024prefpaint} and the inpainting variant of Stable Diffusion 2~\cite{Rombach_2022_CVPR}. We also compare with Difix3D~\cite{wu2025difix3d+}, which applies single-step diffusion to suppress residual artifacts. This setup ensures a fair comparison across compression-only, optimization-based, and generative approaches.

\subsubsection{Implementation Details.}
To preserve fine details under heavy compression while maintaining efficiency, we enhance the VAE decoder with lightweight skip adapters and integrate LoRA modules for parameter-efficient adaptation. During training, the full UNet denoiser is optimized, while only the low-rank adapters on the VAE are updated, balancing flexibility and stability. The overall loss is  
\[
\mathcal{L} = \mathcal{L}_{\text{L2}} + \lambda_{p}\,\mathcal{L}_{\text{LPIPS}} + \lambda_{\text{lat}}\,\mathcal{L}_{\text{latent}},
\]  
where $\lambda_{p}$ and $\lambda_{\text{lat}}$ are weighting factors. UGC and GaussPainter are trained jointly under this unified objective to ensure consistency.  
Further implementation details are provided in Supplement.

\begin{table*}[t]
\centering
\caption{
\textbf{Quantitative results on ScanNet++ under different compression ratios of Gaussian counts, reporting LPIPS$\downarrow$, SSIM$\uparrow$, PSNR$\uparrow$, and compression ratio relative to the source file (404.73 MB).} 
LightGS* indicates the optimized version of LightGS. 
\textbf{Bold} indicates the best performance, and \underline{underline} denotes the second-best.
}
\label{tab:compression_results}
\small
\resizebox{\textwidth}{!}{
\begin{tabular}{l|cccc|cccc|cccc}
\toprule[1.5pt]
\multirow{2}{*}{Method} 
& \multicolumn{4}{c|}{0.1} 
& \multicolumn{4}{c|}{0.2} 
& \multicolumn{4}{c}{0.3} \\
& LPIPS$\downarrow$ & SSIM$\uparrow$ & PSNR$\uparrow$ & Ratio (Size, MB)
& LPIPS$\downarrow$ & SSIM$\uparrow$ & PSNR$\uparrow$ & Ratio (Size, MB)
& LPIPS$\downarrow$ & SSIM$\uparrow$ & PSNR$\uparrow$ & Ratio (Size, MB) \\
\midrule

LightGS (w/o SH)   
& 0.4583 & 0.6130 & 15.56 & 36.5× (11.10)
& 0.3336 & 0.7577 & 19.71 & 18.2× (22.19)
& 0.2667 & 0.8270 & 22.45 & 12.2× (33.29) \\

LightGS (with SH)  
& 0.4539 & 0.6186 & 15.64 & 10.0× (40.47)
& 0.3214 & 0.7750 & 20.20 & 5.0× (80.95)
& \underline{0.2498} & \underline{0.8510} & \underline{23.45} & 3.3× (121.42) \\

LightGS*           
& \underline{0.4062} & \underline{0.6842} & \underline{18.87} & 10.0× (40.47)
& 0.3374 & 0.7393 & \underline{20.59} & 5.0× (80.93)
& 0.2962 & 0.7774 & 21.78 & 3.3× (121.41) \\

FlexGaussian       
& 0.4887 & 0.6833 & 15.62 & \underline{60.3× (6.71)}
& 0.4300 & \underline{0.7899} & 20.09 & \underline{31.3× (12.92)}
& 0.3704 & 0.8408 & 23.23 & \underline{21.2× (19.12)} \\

\midrule

Difix3D            
& 0.4449 & 0.6037 & 15.60 & \multicolumn{1}{c|}{--}
& \underline{0.3188} & 0.7528 & 19.67 & \multicolumn{1}{c|}{--}
& 0.2535 & 0.8228 & 22.33 & \multicolumn{1}{c}{--} \\

PrefPaint          
& 0.5518 & 0.5263 & 13.91 & \multicolumn{1}{c|}{--}
& 0.4547 & 0.6323 & 15.28 & \multicolumn{1}{c|}{--}
& 0.4086 & 0.6837 & 16.76 & \multicolumn{1}{c}{--} \\

SD2-Inpaint        
& 0.5893 & 0.4691 & 12.64 & \multicolumn{1}{c|}{--}
& 0.4988 & 0.5691 & 13.63 & \multicolumn{1}{c|}{--}
& 0.4583 & 0.6130 & 14.07 & \multicolumn{1}{c}{--} \\

\midrule
\rowcolor{gray!20}
\textbf{Ours}      
& \textbf{0.2321} & \textbf{0.8122} & \textbf{22.93} & \textbf{109.1× (3.71)}
& \textbf{0.1948} & \textbf{0.8387} & \textbf{24.29} & \textbf{60.4× (6.70)}
& \textbf{0.1718} & \textbf{0.8603} & \textbf{25.25} & \textbf{37.9× (10.68)} \\

\bottomrule[1.5pt]
\end{tabular}}
\end{table*}

\begin{table}[t]
\centering
\caption{\textbf{Quantitative comparison of different methods under various compression ratios of Gaussian counts on Replica.} We report LPIPS$\downarrow$, SSIM$\uparrow$, PSNR$\uparrow$, and the compression ratio relative to the source file (354.77 MB). LightGS* indicates the optimized version of LightGS. \textbf{Bold} indicates the best performance, and \underline{underline} denotes the second-best. }
\label{tab:replica}
\small
\resizebox{\linewidth}{!}{
\begin{tabular}{l|cccc|cccc|cccc}
\toprule[1.5pt]
\multirow{2}{*}{Method} & \multicolumn{4}{c|}{0.1} & \multicolumn{4}{c|}{0.2} & \multicolumn{4}{c}{0.3} \\
 & LPIPS$\downarrow$ & SSIM$\uparrow$ & PSNR$\uparrow$ & Ratio (Size, MB) 
 & LPIPS$\downarrow$ & SSIM$\uparrow$ & PSNR$\uparrow$ & Ratio (Size, MB) 
 & LPIPS$\downarrow$ & SSIM$\uparrow$ & PSNR$\uparrow$ & Ratio (Size, MB) \\
\midrule
LightGS (w/o SH)  & 0.4158 & 0.7000 & 18.51 & 36.5$\times$ (9.73)  
                  & 0.2645 & 0.8402 & 23.07 & 18.2$\times$ (19.46) 
                  & 0.2130 & 0.8812 & 25.44 & 12.2$\times$ (29.18) \\
LightGS (with SH) & 0.3871 & 0.7150 & 18.93 & 10.0$\times$ (35.48) 
                  & \underline{0.2393} & \underline{0.8564} & 23.94 & 5.0$\times$ (70.95)  
                  & \underline{0.1950} & \textbf{0.8969} & \underline{26.58} & 3.3$\times$ (106.43) \\
LightGS*          & \underline{0.3242} & \underline{0.8122} & \underline{23.79} & 10.0$\times$ (35.47) 
                  & 0.2843 & 0.8421 & \underline{25.34} & 5.0$\times$ (70.93)
                  & 0.2669 & 0.8576 & 25.95 & 3.3$\times$ (106.43) \\
FlexGaussian      & 0.4451 & 0.7237 & 18.91 & \underline{59.0$\times$ (6.01)}  
                  & 0.3496 & 0.8200 & 23.76 & \underline{30.8$\times$ (11.52)} 
                  & 0.3125 & 0.8510 & 26.21 & \underline{20.8$\times$ (17.03)} \\
\midrule
Difix3D           & 0.4264 & 0.6785 & 17.94 & -- 
                  & 0.2821 & 0.8210 & 22.30 & -- 
                  & 0.2358 & 0.8615 & 24.65 & -- \\
PrefPaint         & 0.5201 & 0.5701 & 13.76 & -- 
                  & 0.4112 & 0.6688 & 15.43 & -- 
                  & 0.3776 & 0.6928 & 16.02 & -- \\
SD2-Inpaint       & 0.5451 & 0.5299 & 12.74 & -- 
                  & 0.4404 & 0.6233 & 14.05 & -- 
                  & 0.4073 & 0.6482 & 14.50 & -- \\
\rowcolor{gray!20}
\midrule
Ours              & \textbf{0.2197} & \textbf{0.8298} & \textbf{24.32} & \textbf{107.2$\times$ (3.31)} 
                  & \textbf{0.1652} & \textbf{0.8717} & \textbf{26.81} & \textbf{53.3$\times$ (6.65)} 
                  & \textbf{0.1478} & \underline{0.8844} & \textbf{27.22} & \textbf{35.9$\times$ (9.89)} \\
\bottomrule[1.5pt]
\end{tabular}
}
\end{table}

\subsection{Comparisons}

\subsubsection{Comparison on Indoor Datasets.}
In the indoor datasets ScanNet++\cite{yeshwanthliu2023scannetpp} and Replica\cite{replica19arxiv}, we compare different compression and generation methods under three compression ratios, where the ratios refer to the proportion of Gaussians retained. As shown in Tab.~\ref{tab:compression_results}, our method achieves the best overall performance across all settings. 
On ScanNet++, at the most aggressive ratio (0.1) our approach delivers significant gains in perceptual and structural quality while reaching over $100\times$ compression, clearly surpassing other methods, with a PSNR improvement of about 4\,dB over the second-best baseline. Similar improvements are observed at ratios 0.2 and 0.3, where our method further enhances fidelity while maintaining fast inference speed.
On Replica, our framework consistently outperforms alternatives, sustaining PSNR above 26\,dB and SSIM around 0.88 even at higher compression levels. These results confirm that our approach not only reduces artifacts and structural distortions but also preserves fine details and scene fidelity under extreme compression.
For completeness, detailed rate--distortion curves on ScanNet++ and Replica are provided in the appendix, which visually corroborate the compression trends discussed above.
\subsubsection{Comparison on Outdoor Datasets.}
Tab.~\ref{tab:mipnerf_results} reports results on the Mip-NeRF360~\cite{barron2021mip} benchmark. Our method again surpasses both compression-based and generative baselines across all ratios. Even at an aggressive ratio of 0.03, our method significantly outperforms the second-best approach, achieving a 3.76\,dB gain in PSNR and a 0.06 reduction in LPIPS, while still reaching over \textbf{$300\times$} compression. At moderate ratios (0.05 and 0.1), the framework maintains superior perceptual metrics with much lower distortion, highlighting its robustness to extreme compression. 
Our method also comprehensively outperforms other methods on Tanks \& Temples dataset~\cite{Knapitsch2017}. Due to space limitations, we place the results on the Tanks \& Temples dataset~\cite{Knapitsch2017} in the appendix.

\subsubsection{Compression on Generated 3D Assets.}
To further evaluate real-world applicability to automatically generated 3D content, we conduct an additional experiment on four representative 3D assets synthesized by several world models\cite{hunyuanworld2025tencent,yang2025matrix3d} and converted into Gaussian representations. The intermediate multi-view renderings produced during asset generation are used as reference images, against which we compress each asset with different methods, render all viewpoints, and compute LPIPS, SSIM, and PSNR over the full-view set. As summarized in Tab.~\ref{tab:hunyuan_world_psnr}, ZipGS consistently achieves the best perceptual and distortion metrics at substantially higher compression ratios than strong compression baselines such as FlexGaussian, indicating that our framework transfers well to generated 3D assets and remains well-suited for real-world streaming and deployment scenarios highlighted in the introduction.

\begin{table}[t]
\centering
\footnotesize
\caption{\textbf{Quantitative comparison on generated 3D assets.} We report LPIPS$\downarrow$, SSIM$\uparrow$, PSNR$\uparrow$ (evaluated over all viewpoints), and compression ratio. \textbf{Bold} indicates the best performance, and \underline{underline} denotes the second-best.}
\label{tab:hunyuan_world_psnr}
\begin{tabular}{lcccc}
\toprule[1.5pt]
Method & LPIPS$\downarrow$ & SSIM$\uparrow$ & PSNR$\uparrow$ & Compression Ratio \\
\midrule
LightGS (w/o SH) & 0.4523 & 0.6821 & 19.45 & 38.2× (12.45) \\
LightGS (with SH) & \underline{0.4238} & \underline{0.7145} & \underline{20.12} & 10.5× (45.23) \\
LightGS* & 0.4856 & 0.6654 & 18.93 & 10.5× (45.23) \\
FlexGaussian & 0.4387 & 0.6987 & 19.78 & \underline{55.3× (9.87)} \\
\midrule
Difix3D & 0.4652 & 0.6512 & 18.34 & -- \\
PrefPaint & 0.5723 & 0.5432 & 15.67 & -- \\
SD2-Inpaint & 0.5891 & 0.5124 & 14.89 & -- \\
\rowcolor{gray!20}
\textbf{Ours} & \textbf{0.3124} & \textbf{0.7823} & \textbf{22.56} & \textbf{98.7× (5.61)} \\
\bottomrule[1.5pt]
\end{tabular}
\end{table}

\begin{table*}[t]
\centering
\caption{\textbf{Quantitative comparison of novel view synthesis on Mip-NeRF360}. We report LPIPS$\downarrow$, SSIM$\uparrow$, PSNR$\uparrow$, and compression ratio relative to the source file (658.91MB). LightGS* indicates the optimized version of LightGS. \textbf{Bold} indicates the best performance, and \underline{underline} denotes the second-best.}
\label{tab:mipnerf_results}
\small
\resizebox{\textwidth}{!}{
\begin{tabular}{lcccc|cccc|cccc}
\toprule[1.5pt]
\multirow{2}{*}{Method} 
& \multicolumn{4}{c|}{0.03} & \multicolumn{4}{c|}{0.05} & \multicolumn{4}{c}{0.1} \\
\cmidrule(lr){2-5}\cmidrule(lr){6-9}\cmidrule(lr){10-13}
& LPIPS$\downarrow$ & SSIM$\uparrow$ & PSNR$\uparrow$ & Ratio (Size, MB)
& LPIPS$\downarrow$ & SSIM$\uparrow$ & PSNR$\uparrow$ & Ratio (Size, MB)
& LPIPS$\downarrow$ & SSIM$\uparrow$ & PSNR$\uparrow$ & Ratio (Size, MB) \\
\midrule[1.5pt]
LightGS (w/o SH)   & 0.5868 & 0.4481 & 15.71 & 121.6× (5.42)
                   & 0.5313 & 0.5072 & 16.72 & 72.9× (9.03)
                   & 0.4585 & 0.5936 & 17.73 & 36.5× (18.07) \\
LightGS (with SH)  & 0.5802 & \underline{0.4605} & 15.76 & 33.3× (19.77)
                   & 0.5224 & \underline{0.5253} & 16.83 & 20.0× (32.95)
                   & 0.4466 & \textbf{0.6187} & 17.99 & 10.0× (65.89) \\
LightGS*           & 0.6840 & 0.4104 & 15.40 & 33.3× (19.77)
                   & 0.6286 & 0.4487 & 15.95 & 20.0× (32.95)
                   & 0.5491 & 0.5134 & 16.82 & 10.0× (65.89) \\
FlexGaussian       & 0.5626 & 0.4522 & 14.81 & \underline{189.9× (3.47)}
                   & 0.5182 & 0.5035 & 15.91 & \underline{120.0× (5.49)}
                   & 0.4464 & 0.5968 & 17.97 & \underline{62.27× (10.58)} \\
\midrule
Difix3D            & \underline{0.5217} & 0.4402 & \underline{15.94} & --
                   & \underline{0.4690} & 0.4973 & \underline{16.97} & --
                   & \underline{0.3999} & 0.5885 & \underline{18.74} & -- \\
PrefPaint          & 0.6574 & 0.3619 & 14.27 & --
                   & 0.6081 & 0.3983 & 14.84 & --
                   & 0.5670 & 0.4135 & 14.73 & -- \\
SD2-Inpaint        & 0.6623 & 0.3428 & 12.95 & --
                   & 0.6248 & 0.3716 & 13.46 & --
                   & 0.5851 & 0.3930 & 13.39 & -- \\
\rowcolor{gray!20}
\textbf{Ours}      & \textbf{0.4595} & \textbf{0.4832} & \textbf{19.70} & \textbf{352.3× (1.87)}
                   & \textbf{0.3934} & \textbf{0.5375} & \textbf{21.01} & \textbf{227.2× (2.90)}
                   & \textbf{0.3303} & \underline{0.6054} & \textbf{22.13} & \textbf{102.15× (6.45)} \\
\bottomrule[1.5pt]
\end{tabular}}
\end{table*}
\begin{table*}[t]
\vspace{-0.2cm}
\centering
\caption{\textbf{Ablation studies of the proposed framework on ScanNet++~\cite{yeshwanthliu2023scannetpp}}. 
(a) Effect of UGC and GaussPainter. 
(b) Effect of mask guidance and latent supervision. 
(c) Inference time comparison on an NVIDIA A100 GPU. }
\vspace{-0.2cm}
\label{tab:ablation_all}
\small
\begin{subtable}{0.48\linewidth}
\centering
\caption{Effect of pipeline components}
\vspace{-0.2cm}
\begin{tabular}{ccccc}
\toprule[1.5pt]
UGC & GaussPainter & LPIPS$\downarrow$ & SSIM$\uparrow$ & PSNR$\uparrow$ \\
\midrule
$\times$ & $\times$ & 0.3336 & 0.7577 & 19.71 \\
\checkmark & $\times$ & 0.2985 & 0.7794 & 21.73 \\
$\times$ & \checkmark & 0.2059 & 0.8336 & 23.24 \\
\rowcolor{gray!20}
\checkmark & \checkmark & \textbf{0.1948} & \textbf{0.8387} & \textbf{24.29} \\
\bottomrule[1.5pt]
\end{tabular}
\label{tab:ablation_pipeline}
\end{subtable}
\hfill
\begin{subtable}{0.48\linewidth}
\centering
\caption{Inference time (A100)}
\vspace{-0.2cm}
\begin{tabular}{lc}
\toprule[1.5pt]
Method & Time (ms) \\
\midrule
Difix3D & \underline{230.36 $\pm$ 16.47} \\
PrefPaint & 2724.02 $\pm$ 24.32 \\
SD2-Inpaint & 2683.14 $\pm$ 61.76 \\
\rowcolor{gray!20}
\textbf{Ours} & \textbf{65.85 $\pm$ 0.63} \\
\bottomrule[1.5pt]
\end{tabular}
\label{tab:inference_time}
\end{subtable}

\begin{subtable}{0.44\linewidth}
\centering
\caption{Effect of design choices}
\vspace{-0.2cm}
\begin{tabular}{lccc}
\toprule[1.5pt]
Method & LPIPS$\downarrow$ & SSIM$\uparrow$ & PSNR$\uparrow$ \\
\midrule
Baseline & 0.3464 & 0.7480 & 21.05 \\
+ Mask guidance & 0.3305 & 0.7676 & 21.73 \\
\rowcolor{gray!20}
+ Latent supervision & \textbf{0.2321} & \textbf{0.8122} & \textbf{22.93} \\
\bottomrule[1.5pt]
\end{tabular}
\label{tab:ablation_latent}
\end{subtable}
\hfill
\end{table*}
\subsection{Ablation Studies}
\subsubsection{Effect of Pipeline Modules.}  
To evaluate the impact of the compression stage, we conduct ablations using only Global Significance Score pruning as the baseline. 
As shown in Tab.~\ref{tab:ablation_pipeline}, adding UGC on top of this baseline improves structural similarity and reconstruction fidelity, while further integrating GaussPainter enhances perceptual quality. 
When the two modules are combined, the system achieves the best overall performance, reducing LPIPS to 0.1948 and increasing SSIM/PSNR to 0.8387 and 24.29 dB. 
This experiment is conducted on ScanNet++~\cite{yeshwanthliu2023scannetpp} with a 20\% pruning ratio.  
Since GaussPainter is the core generation module, we further analyze its internal components: mask guidance and latent supervision (Tab.~\ref{tab:ablation_latent}). This ablation is carried out on the ScanNet++ test set under a 10\% compression ratio. Incorporating mask guidance improves perceptual quality by helping the model focus on relevant regions, while adding latent supervision yields the most significant gains, demonstrating that latent-level consistency provides a stronger learning signal and substantially improves perceptual fidelity.


\begin{table}[t]
\centering
\caption{\textbf{Compatibility with different compression methods on ScanNet++}. Results are reported at 30\% Gaussian retention using PSNR$\uparrow$, LPIPS$\downarrow$, and SSIM$\uparrow$. Adding GaussPainter consistently improves both LightGS and MaskGaussian.}
\label{tab:compatibility}
\resizebox{\textwidth}{!}{
\small
\begin{tabular}{lccc|lccc}
\toprule[1.5pt]
Method & PSNR $\uparrow$ & LPIPS $\downarrow$ & SSIM $\uparrow$
& Method & PSNR $\uparrow$ & LPIPS $\downarrow$ & SSIM $\uparrow$ \\
\midrule[1.5pt]
LightGS & 27.55 & 0.1615 & 0.9039 
& MaskGaussian & 25.87 & 0.3078 & 0.8198 \\
LightGS + GaussPainter & \textbf{28.31} & \textbf{0.1271} & \textbf{0.9164} 
& MaskGaussian + GaussPainter & \textbf{26.56} & \textbf{0.1340} & \textbf{0.8770} \\
\bottomrule[1.5pt]
\end{tabular}
}
\vspace{-0.2cm}
\end{table}
\subsubsection{Compatibility with Other Compression Methods.}
To evaluate the generality of GaussPainter, we apply it to scenes compressed by LightGS~\cite{fan2024lightgaussian} and MaskGaussian~\cite{liu2025maskgaussian} on ScanNet++ at 30\% Gaussian retention. As shown in Tab.~\ref{tab:compatibility}, GaussPainter consistently improves all metrics across different pruning backbones, demonstrating that it is backbone-agnostic and can serve as a plug-and-play refinement module for existing Gaussian compression strategies.

\section{Conclusion}
In this work, we introduced ZipGS, a feed-forward framework for extreme compression of 3D Gaussian representations that reconciles aggressive pruning with high-fidelity rendering. By decoupling compression and restoration into Universal Gaussian Compression (UGC) and GaussPainter, ZipGS removes redundant Gaussians without per-scene optimization and leverages mask-guided, latent-space diffusion to recover missing structures while refining visible regions. Experiments on ScanNet++, Replica, Mip-NeRF360, Tanks \& Temples, and automatically generated 3D assets show that ZipGS consistently outperforms strong compression and generative baselines across a wide range of ratios while remaining compatible with existing pruning backbones, highlighting diffusion priors as a practical tool for streaming and deploying Internet-scale 3D Gaussian assets.
\clearpage  

%
%
\bibliographystyle{splncs04}
\bibliography{main}
\clearpage

\section{Supplement}
\subsection{Training Details}
\label{app:training_details}

All experiments are conducted on NVIDIA A100 GPUs with mixed-precision training. We train for up to 10{,}000 steps with a learning rate of $2\times10^{-5}$. The backbone components—tokenizer, text encoder, VAE, and U\!-\!Net—are initialized from the pre-trained \texttt{sd-turbo} model~\cite{Rombach_2022_CVPR,radford2021learning}. We adopt a one-step diffusion scheduler implemented via \texttt{DDPMScheduler} and apply mask conditioning to guide the denoising process.

During training, \emph{UGC} and \emph{GaussPainter} are jointly optimized with a per-GPU batch size of 1. Each forward pass processes a single scene: Gaussian primitives are compressed by \emph{UGC}, and a randomly sampled view is rendered into an image that \emph{GaussPainter} uses for reconstruction.

The variational autoencoder (VAE) is augmented with LoRA (rank 4). LoRA adapters are inserted into all \texttt{Conv2d} layers of the VAE decoder, with Gaussian-initialized LoRA weights to ensure stable optimization.
\subsection{Additional Experimental Results}
This section complements the main paper with results under additional compression ratios, extended qualitative visualizations, and further ablations.



\subsubsection{Indoor benchmarks at 20\% and 30\% ratios.}
\begin{figure*}[t]
  \centering
  \includegraphics[width=\textwidth]{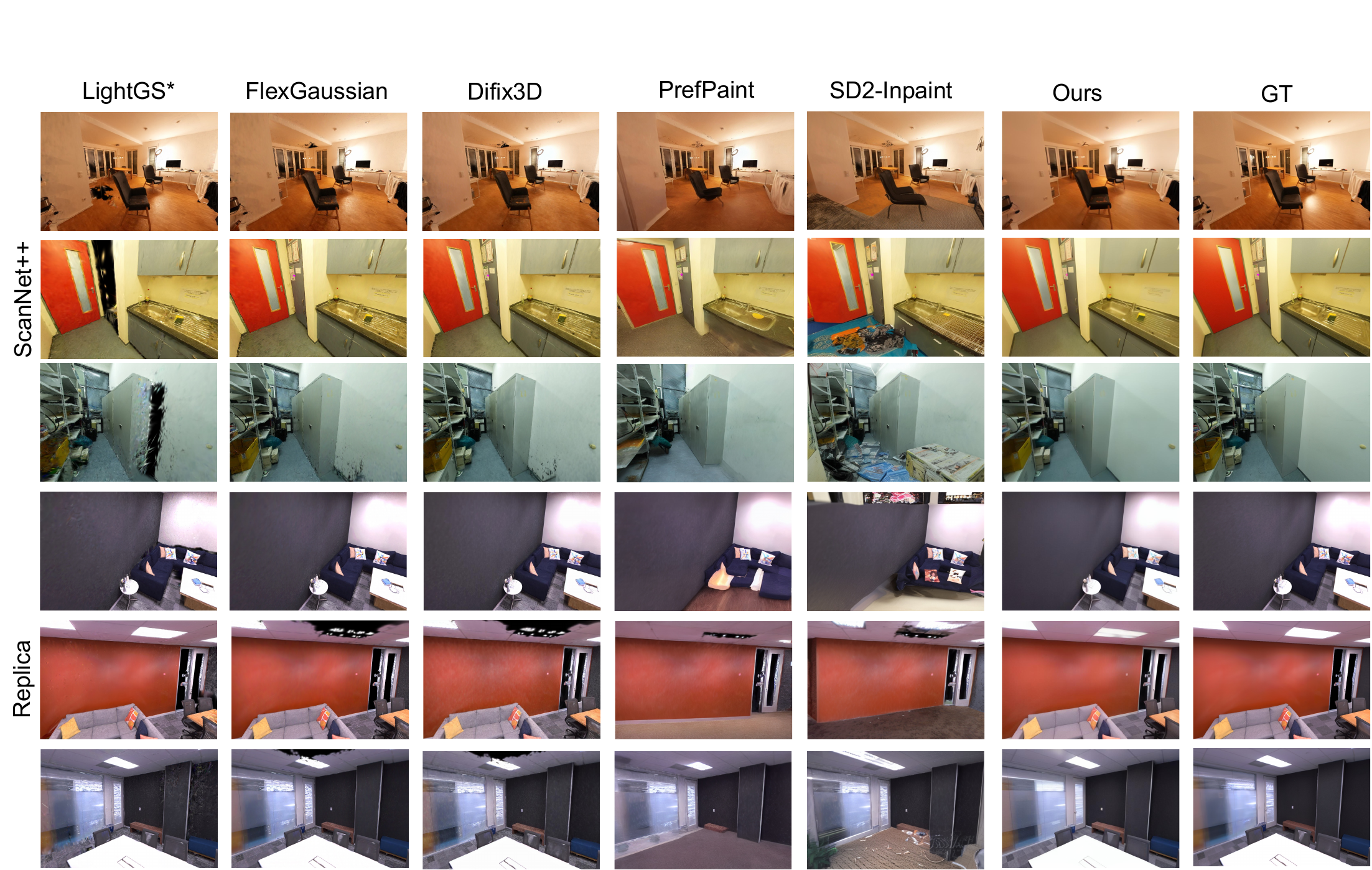}%
  \caption{\textbf{Indoor results at $r{=}0.20$ (20\% pruning).} Qualitative comparisons on ScanNet++ and Replica. Here $r$ denotes the fraction of Gaussians retained after compression; for example, $r{=}0.20$ means 20\% of the original Gaussians are preserved. For each scene we show (left to right): compressed input rendering, our reconstruction, and zoom-in patches. Our method preserves structure while recovering fine textures.}
  \label{fig:indoor_20}
\end{figure*}

\begin{figure*}[t]
  \centering
  \includegraphics[width=\textwidth]{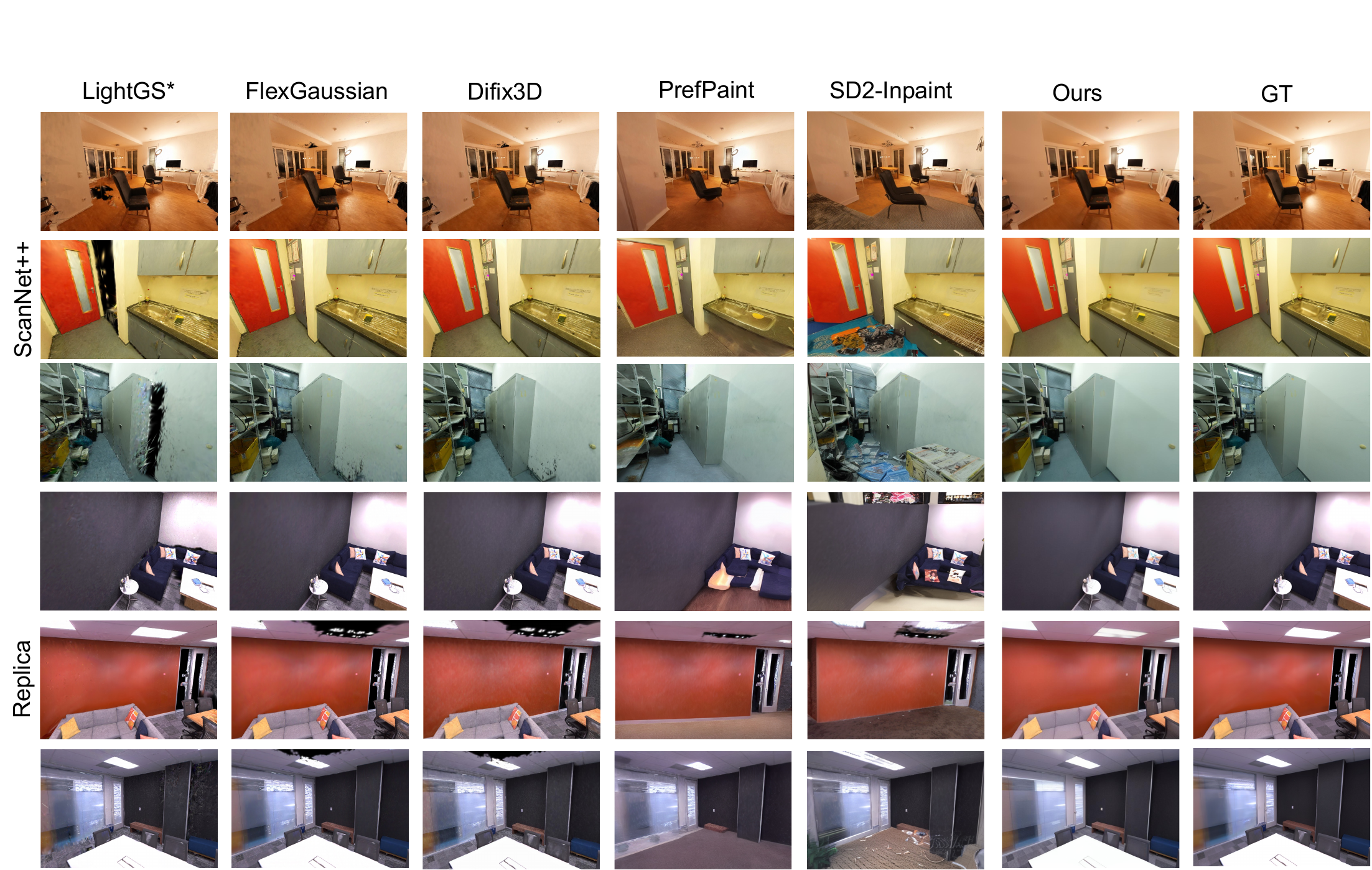}%
  \caption{\textbf{Indoor results at $r{=}0.30$ (30\% pruning).} Additional qualitative comparisons on ScanNet++ and Replica. 
  Here $r{=}0.30$ means that 30\% of the original Gaussians are retained. 
  Even under this stronger compression, the proposed pipeline mitigates artifacts and maintains perceptual quality.}
  \label{fig:indoor_30}
\end{figure*}

Fig.~\ref{fig:indoor_20} and \ref{fig:indoor_30} present qualitative comparisons on \textbf{ScanNet++} and \textbf{Replica} at pruning ratios $r{=}0.20$ and $r{=}0.30$, respectively. For each scene, we show the input rendering after compression, our reconstruction, and zoom-in crops. These visualizations corroborate the quantitative trends reported in the main paper: as the pruning ratio increases, UGC preserves structural fidelity while GaussPainter effectively restores texture and color consistency, yielding perceptually sharper details and fewer artifacts.

\subsubsection{Outdoor benchmarks at 3\% and 5\% ratios.}
\begin{figure*}[t]
  \centering
  \includegraphics[width=\textwidth]{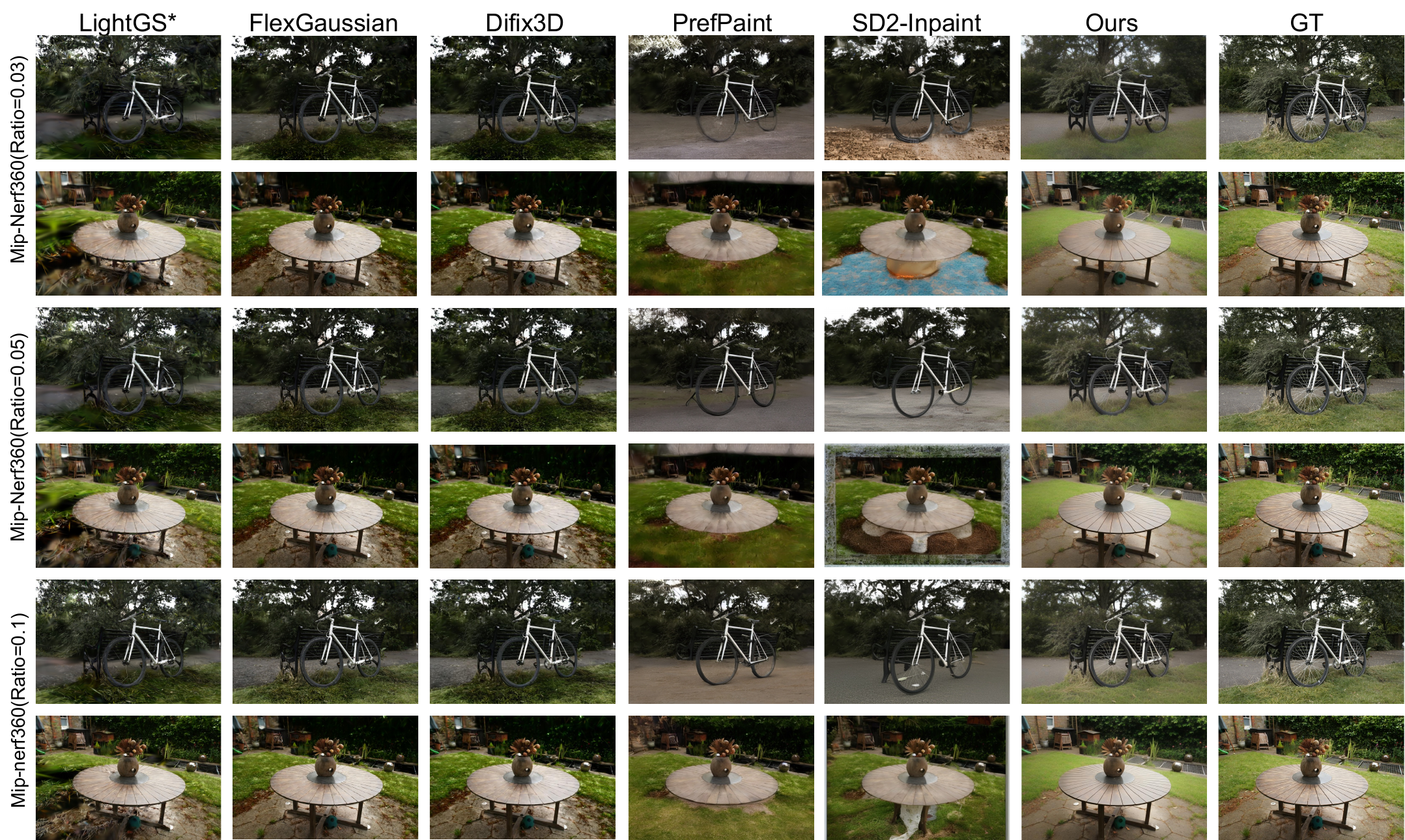}%
  \caption{\textbf{Outdoor results on Mip-NeRF360 at $r{=}0.03$ and $r{=}0.05$.} 
  Here $r{=}0.03$ and $r{=}0.05$ indicate that only 3\% and 5\% of the original Gaussians are retained, respectively. 
  Despite such extreme compression, our approach preserves coherent geometry and reduces color drift, particularly in low-texture or large-hole regions.}
  \label{fig:mip_outdoor}
\end{figure*}

Fig.~\ref{fig:mip_outdoor} shows additional results on Mip-NeRF360 at extreme ratios $r{=}0.03$ and $r{=}0.05$. Despite aggressive compression, the combination of mask-guided conditioning and one-step diffusion maintains coherent geometry and suppresses color drift in low-texture regions, demonstrating robustness in challenging outdoor scenes.

\begin{figure*}[t]
  \centering
  \includegraphics[width=\textwidth]{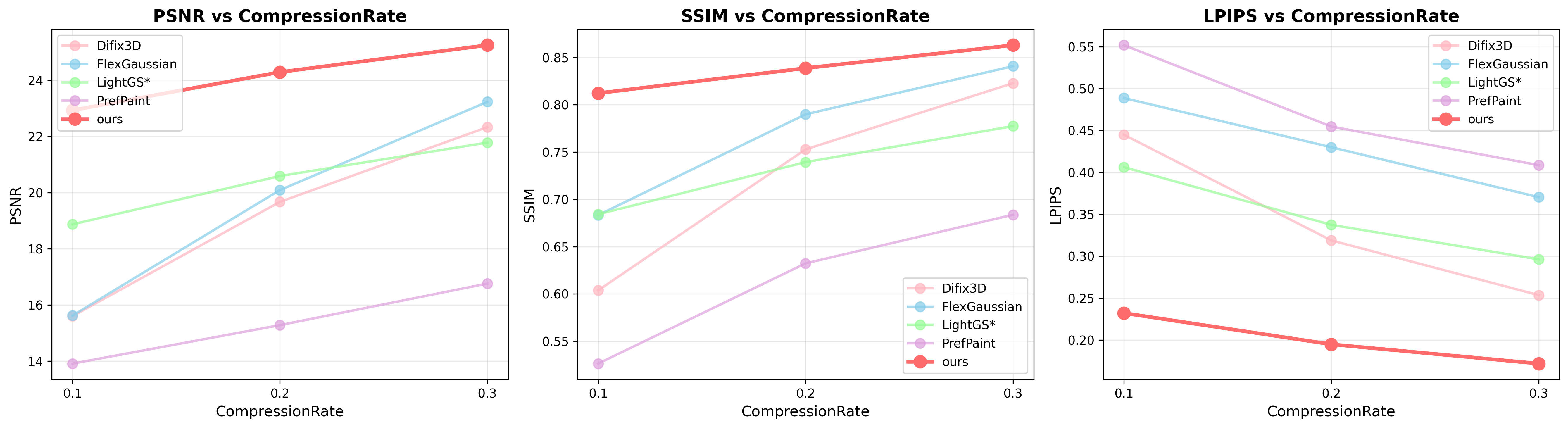}%
  \caption{\textbf{Quantitative results on ScanNet++ under different compression ratios of Gaussian counts, reporting LPIPS$\downarrow$, SSIM$\uparrow$, PSNR$\uparrow$.}}
  \label{fig:scannet_compare}
\end{figure*}

\begin{figure*}[t]
  \centering
  \includegraphics[width=\textwidth]{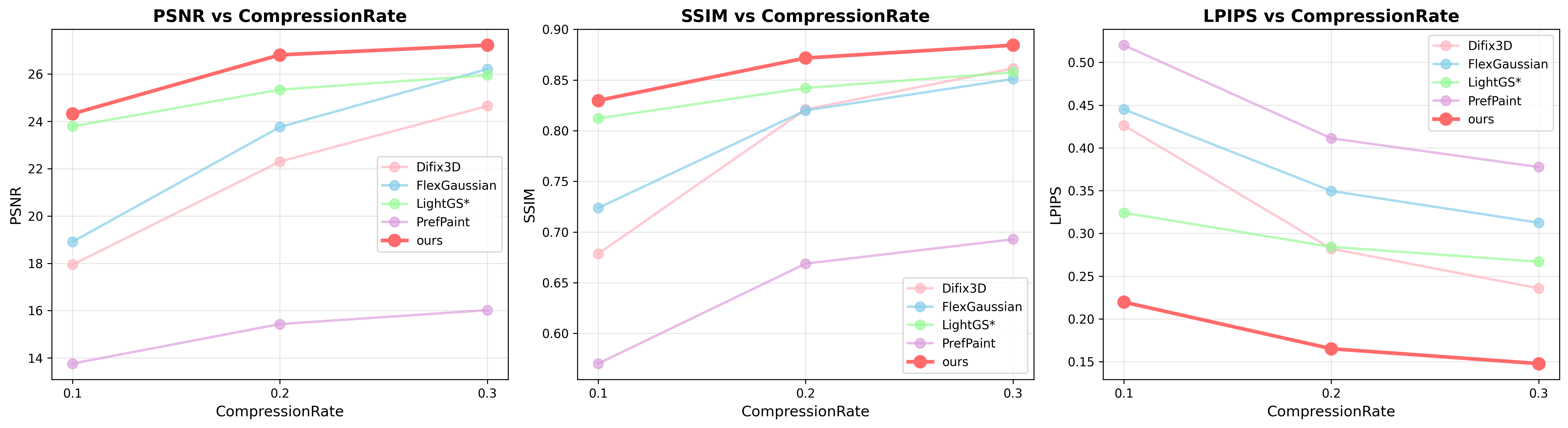}%
  \caption{\textbf{Quantitative results on Replica under different compression ratios of Gaussian counts, reporting LPIPS$\downarrow$, SSIM$\uparrow$, PSNR$\uparrow$.}}
  \label{fig:replica_compare}
\end{figure*}

\begin{figure*}[t]
  \centering
  \includegraphics[width=\textwidth]{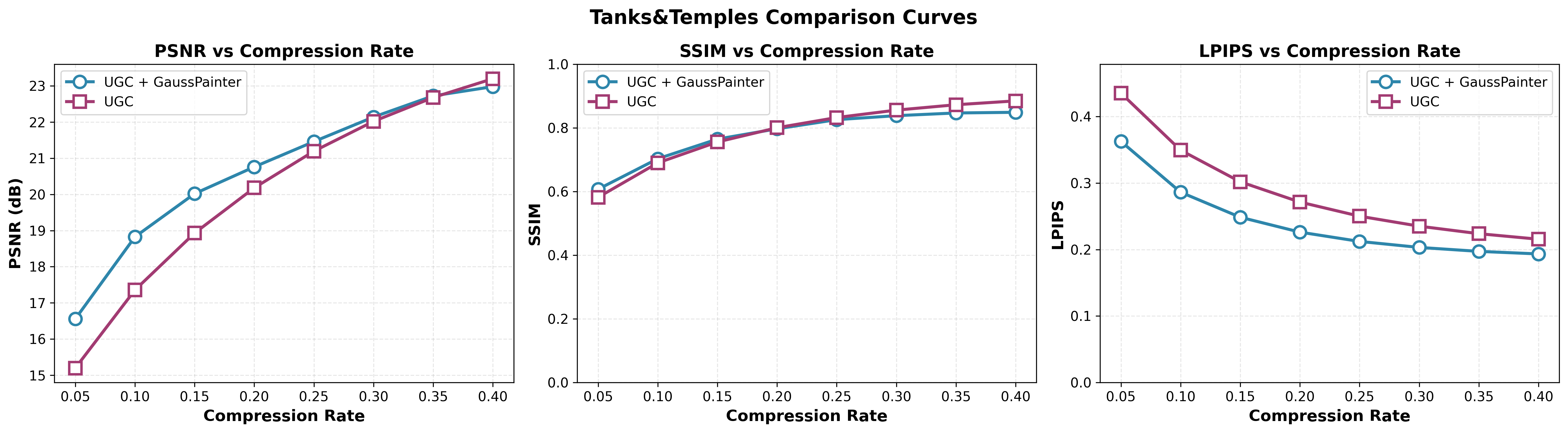}%
  \caption{\textbf{Outdoor results on Tanks \& Temples datasets} 
  This figure presents three curves (PSNR, SSIM, LPIPS) of UGC + GaussPainter and UGC under different compression rates on the unseen scenes of Tanks \& Temples dataset. 
  Across metrics, UGC + GaussPainter performs better or comparable to UGC in most cases—its gains are prominent under aggressive compression.}
  \label{fig:tank&temple}
\end{figure*}

\begin{figure*}[t]
  \centering
  \includegraphics[width=\textwidth]{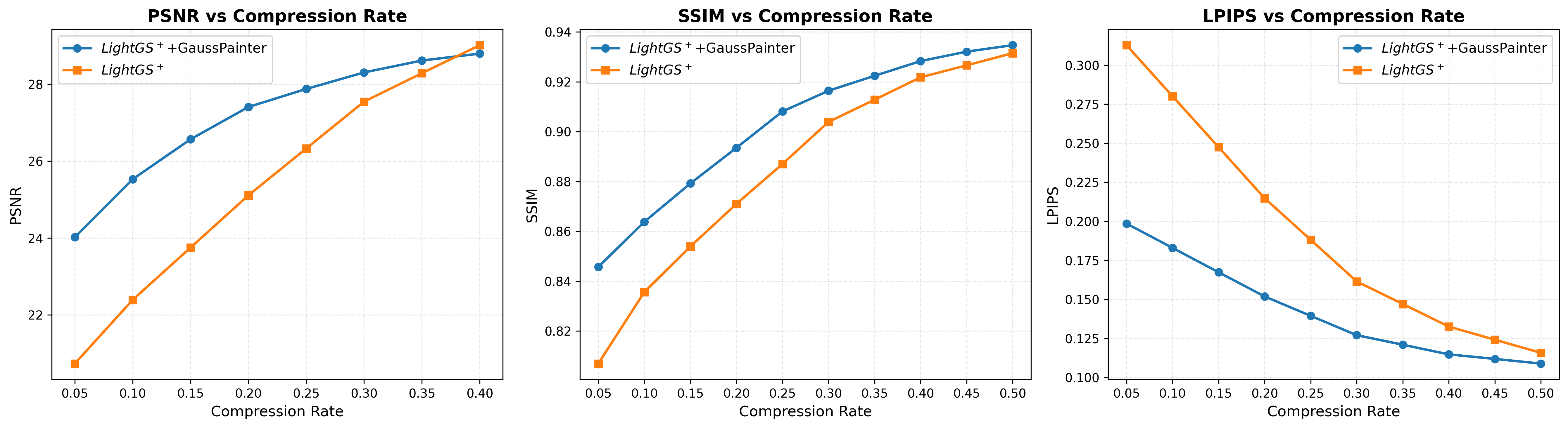}%
  \caption{\textbf{Quantitative results on ScanNet++ under different compression ratios of Gaussian counts, comparing the performance of LightGS and LightGS+GaussPainter across PSNR, SSIM, and LPIPS metrics.}}
  \label{fig:compression_metrics_comparison}
\end{figure*}
\subsection{Details of Data Sources and Training Details}
The Gaussian scenes used in our experiments are derived from publicly available datasets. For indoor benchmarks, ScanNet++ and Replica Gaussian scenes are obtained from the SceneSplat repository~\cite{li2025scenesplat}. For outdoor scenes on Mip-NeRF360 \cite{barron2021mip} and DL3DV \cite{ling2024dl3dv}, Gaussian scenes are generated using the 3DGS pipeline with 30{,}000 optimization steps; densification and other hyperparameters follow the default 3DGS settings unless otherwise stated.

We train our models using two data sources: DL3DV and ScanNet++. The training set includes all video sequences from DL3DV and the official training split of ScanNet++. For validation, we randomly render 100 novel views from the training scenes. For testing, we use the 50 official test scenes from ScanNet++ and all scenes from Replica and Mip-NeRF360.

\subsection{Generalization to Unseen Scenes and Performance Trends.}
To further evaluate how well our method generalizes beyond the training distribution, we conduct experiments on two scenes from the Tanks \& Temples dataset, the Mip-NeRF360 dataset, and the Replica dataset. All these scenes are not included in our training data, ensuring that the model has never observed any images from these environments, and thus Tanks \& Temples, Mip-NeRF360, and Replica all serve as out-of-domain data for testing generalization performance.

The Fig. \ref{fig:tank&temple} reports the PSNR, SSIM, and LPIPS curves under different compression rates. Across the three metrics, UGC + GaussPainter performs consistently better or comparable to UGC across most compression levels. The gains are most notable under aggressive compression, where the Gaussian representation becomes extremely sparse and GaussPainter effectively recovers missing structures. At higher compression budgets, UGC alone occasionally achieves slightly better scores; we believe this is because the Tanks \& Temples scenes contain complex outdoor geometry and fine textures, where the generative refinement may introduce subtle artifacts when the underlying Gaussian representation is already of high quality. As summarized in Tab.~\ref{tab:replica}, the Replica results are reported in this table. The quantitative comparisons on the Mip-NeRF360 dataset are reported in Tab.~\ref{tab:mipnerf_results}.

Overall, these results show that our approach generalizes well to unseen scenes and provides clear benefits under challenging compression settings, while maintaining competitive performance in high-quality regimes.

\end{document}